\documentclass[11pt,a4paper]{article}
\usepackage[hyperref]{emnlp2020}
\usepackage{times}
\usepackage{latexsym}

\usepackage{microtype}

\aclfinalcopy 

\title{Efficient Inference For Neural Machine Translation}

\author{Yi-Te Hsu$^{1}$\thanks{~~~~Work done during internship at Apple Inc.} \qquad
  Sarthak Garg$^{2}$ \qquad
  Yi-Hsiu Liao$^{2}$ \qquad
  Ilya Chatsviorkin$^{2}$ \\
  $^{1}$Johns Hopkins University\\
  $^{2}$Apple Inc.\\
  \texttt{yhsu16@jhu.edu, \{sarthak\_garg, yihsiu\_liao, ilych\}@apple.com}\\}

\date{}

\usepackage{multirow} 
\usepackage{graphicx}
\usepackage{booktabs}
\usepackage{amsmath}

\begin{document}
\maketitle
\begin{abstract}
Large Transformer models have achieved state-of-the-art results in neural machine translation and have become standard in the field.
In this work, we look for the optimal combination of known techniques to optimize inference speed without sacrificing translation quality. We conduct an empirical study that stacks various approaches and demonstrates that combination of replacing decoder self-attention with simplified recurrent units, adopting a deep encoder and a shallow decoder architecture and multi-head attention pruning can achieve up to $109$\% and $84$\% speedup on CPU and GPU respectively and reduce the number of parameters by $25$\% while maintaining the same translation quality in terms of BLEU.
\end{abstract}

\section{Introduction and Related Work}
\label{sec:introductionrelatedwork}
Transformer models \citep{Attention-is-All-You-Need} have outperformed previously used RNN models and traditional statistical MT techniques.  This improvement, though, comes at the cost of higher computation complexity. The decoder computation often remains the bottleneck due to its autoregressive nature, large depth and self-attention structure. 

There has been a recent trend towards making the models larger and ensembling multiple models to achieve the best possible translation quality \citep{GShardSG_google_Lepikhin2020GShardSG, Efficient_google_NIPS2019_8305}. Leading solutions on common benchmarks \citep{NewMT:Zhu2020Incorporating, gpt3:brown2020language} usually use an ensemble of Transformer big models, which combined can have more than $1$ billion parameters. 

Previous works suggest replacing the expensive self-attention layer in the decoder with simpler alternatives like the Average Attention Network (AAN) \citep{AAN-zhang-Etal:2018}, Simple Recurrent Unit (SRU) \citep{SRU:lei-etal-2018-simple} and Simpler Simple Recurrent Unit (SSRU) \citep{ludicrously:kim2019}. AAN is a simpler version of the self-attention layer which places equal attention weights on all previously decoded words instead of dynamically computing them. SRU and SSRU are lightweight recurrent networks, with SSRU consisting of only $2$ matrix multiplications per decoded token.

Because of the autoregressive property of the decoder in a standard Transformer model, reducing computation cost in the decoder is much more important than in the encoder. Recent publications \citep{deep:miceli-barone-etal-2017-deep, deep:wang-etal-2019-learning, deepencoder} thus suggest that a deep encoder, shallow decoder architecture can speed up inference while maintaining a similar BLEU score.

Another line of research focuses on model pruning techniques to make NMT models smaller and more efficient.
In this paper, we only explore structured pruning methods, in which smaller components of the network are pruned away.
 Applications of structured pruning to NMT include works by \citet{prune_voita-etal-2019-analyzing} and \citet{prune_michel:NIPS2019_9551} which show that most of the attention heads in the network learn redundant information and can be pruned. \citet{prune_michel:NIPS2019_9551} proposed the idea of pruning heads by head importance scoring. \citet{prune_voita-etal-2019-analyzing} uses a relaxation of $L_0$ regularization \citep{L0:louizos2018learning} to prune the attention heads.

All of the above mentioned methods 
use the vanilla Transformer architecture as their baseline, so it is not clear if these approaches can give complimentary results when combined together. In this work, we explore and benchmark, combining all of the above techniques, with the goal of maximizing inference speed without hurting translation quality.

After carefully stacking the approaches, our proposed architecture is able to achieve a significant speed improvement of $84$\% on GPU and $109$\% on CPU architectures without any degradation of translation quality in terms of BLEU.

\section{Efficient Inference for Neural Machine Translation}
\label{sec:method}
This section presents the proposed efficient inference architecture for neural machine translation. First, we outline the overall procedure of building an efficient inference architecture. Then, we detail each step in the process.


\begin{figure}[ht]
    \centering
    \includegraphics[width=0.5\textwidth]{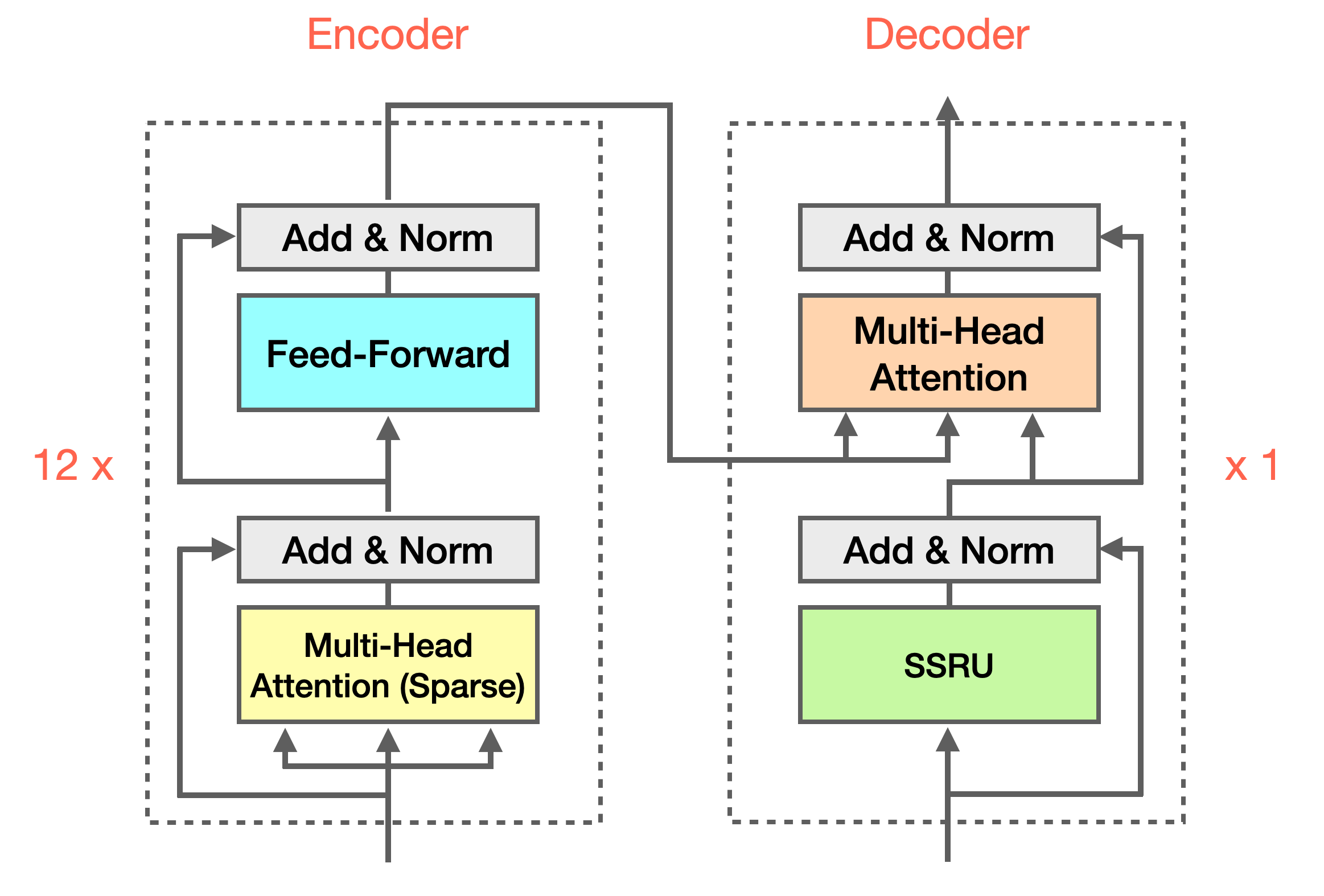}
    \caption{\label{figure:efficient_transformer} Efficient Transformer Architecture}
\end{figure}

First, we use sequence-level knowledge distillation \citep{KD-kim-rush-2016-sequence} to transfer knowledge from a strong teacher model to a smaller student model. This approach allows the student model to learn from a simpler target distribution and therefore enables us to use a simpler architecture.

Then, to simplify the decoder of the student model, the self-attention mechanism is replaced by lightweight recurrent units \citep{ludicrously:kim2019}, and the feed-forward network is removed. To further reduce the decoder computation, we adopt the deep encoder, shallow decoder architecture \citep{deepencoder}. 
Lastly, we prune redundant attention heads through $L_0$ regularization \citep{prune_voita-etal-2019-analyzing}. Each architecture modification is performed by retraining the student model.
Figure~\ref{figure:efficient_transformer} shows the proposed efficient Transformer architecture. 

\subsection{Teacher-student Training}
\label{ssec:teacherstudent}
We follow the procedure described in \citet{ludicrously:kim2019}, to train an ensemble of $8$ Transformer-big models, $4$ forward, $4$ reverse direction, as the first round of teacher models (T). 
Without the help of extra monolingual corpora, we apply multi-agent dual learning (MADL) \citep{madl:wang2018multiagent} to train another $8$ Transformer big teacher models (T-MADL) by re-decoded bitext with ensemble teacher models (T) in both directions.
Then we use noisy backward-forward translation \citep{noisy_translation:edunov2018understanding} with the T-MADL model to, again, re-decode the original bitext, but with more variance on the source side.
Finally, we use the above generated synthetic data along with the original bitext to train our student model.

We use interpolated sequence-level knowledge distillation \citep{KD-kim-rush-2016-sequence} in most of the described re-decoding runs except the noisy backward-forward translation where sampling is used in the reverse direction.
More details about model training and architecture can be found in \citet{ludicrously:kim2019}. 

\subsection{Replacing Self-attention with Lightweight Recurrent Units}
\label{ssec:lightweightRU}
Inspired by \citet{ludicrously:kim2019}, we replace the decoder self-attention with an RNN, reducing its
time complexity from $O(N^2)$ to $O(N)$, where $N$ is the length of the output sentence. 
We compare replacing self-attention with two lightweight layers: SSRU and AAN, in Section \ref{ssec:ssru_result}. The SSRU layer is as follows:
\begin{equation}\label{eq:ssru}
\begin{split}
f_t &= \sigma(W_t x_t + b_f) \\
c_t &= f_t \odot c_{t-1} + (1-f_t) \odot W x_t\\
o_t &= ReLU(c_t)
\end{split}
\end{equation}
where the $\odot$ is element-wise multiplication. $x_t$, $o_t$, $f_{t}$ and $c_{t}$ are the input, output, forget-gate and cell-state, respectively.
We optimized the SSRU by combining the two matrix multiplications, $W_tx_t$ and $Wx_t$, into one. We find this simple trick can improve speed by $6$\% on GPU.

For AAN, we found that removing the gating layer does not degrade the translation quality while reducing the computation. In our experiments, we use the following implementation of AAN (without a gating layer):
\begin{equation}\label{eq:aan}
o_t = FFN(\frac{1}{t}\sum_{k=1}^{t} \textbf{x}_k)
\end{equation}
where $FFN(\cdot)$ is a position-wise two-layer feed-forward network. $t$, $o_t$ and $x_k$ denote the current position, output at position $t$ and input at position $k$ respectively. 

\subsection{Removing the Feed-forward Layer}
\label{ssec:ffn}
Each decoder layer consists of a lightweight recurrent unit, followed by an encoder-decoder multi-head attention component and a pointwise feed-forward layer. 
The feed-forward sub-layer is responsible for $33$\% of parameters within the 6-layer decoder; however, we found that it can be removed entirely from the decoder without hurting the translation quality with our implementation of SSRU (Section \ref{ssec:ssru_result}). 


\subsection{Deep Encoder, Shallow Decoder}
\label{ssec:deepencoder}
In order to further reduce the decoder computation, we decrease the number of decoder layers. In line with the work done by \citet{deepencoder}, to maintain the same model capacity, we increase the number of encoder layers. We explore the speed-accuracy trade-off while varying the depth of both components in Section \ref{ssec:layer_result}, and find that using $12$ encoder layers and $1$ decoder layer gives a significant speedup without losing translation quality.

\subsection{Pruning Attention Heads}
\label{ssec:pruning}
Adopting a deep encoder, shallow decoder architecture achieves a good speed-quality tradeoff; however, it increases the number of parameters in the encoder. 
To further improve efficiency and reduce parameters, we apply multi-head attention pruning proposed by \citet{prune_voita-etal-2019-analyzing} to our architecture.
The output of each head $h_i$ across all
attention layers
is multiplied by a learnable gate $g_i$, before it is passed to subsequent layers of the network. 
To switch off less informative heads (i.e. $g_i = 0$), we applied $L_0$ regularization to the gates. $L_0$ norm is the number of non-zero gates across the model. However, because of the non-differentiable property of the $L_0$ norm, a differentiable approximation is used. 
Each gate $g_i$, is modeled as a random variable sampled from a Hard Concrete Distribution \citep{L0:louizos2018learning} parameterized by $\phi_i$, and takes values in the range $[0,1]$. 
We then minimize the differentiable approximation of $L_0$ regularization loss, $L_c$:
\begin{equation}\label{eq:l0regularization}
L_c(\phi) = \sum_{i=1}^h (1-P(g_i = 0 | \phi_i)),
\end{equation}
where $h$ denotes the total number of heads, $\phi$ is the set of gate parameters, and $P(g_i = 0 | \phi_i)$ is computed according to the Hard Concrete Distribution.

The model is initially trained with the standard cross entropy loss $L_{x_{ent}}$ and then fine-tuned with the additional regularization loss as follows:
\begin{equation}\label{eq:l0regularizationtotalloss}
L(\theta, \phi) = L_{xent}(\theta, \phi) + \lambda L_{c}(\phi),
\end{equation}
where $\theta$ denotes the set of original model parameters, and $\lambda$ is a hyperparameter which controls how aggressively the attention heads are pruned. During inference time, all heads $h_j$, where $P(g_j = 0 | \phi_j) = 1$ are completely removed from the network. Our experiments in Section \ref{ssec:prunehead_results} show that we can effectively prune out a large portion of redundant self-attention heads from the deep-encoder. 

\section{Experiments}
\label{sec:experiment}
We use the Transformer base model \citep{Attention-is-All-You-Need} trained on teacher decoded data as our baseline. All the described methods are stacked on top of this baseline model. 
Following \citet{ludicrously:kim2019}, we use 4 million bitext from the  WMT'14 English-German news translation task. All sentences are encoded with 32K subword units using SentencePiece \citep{sentencePiece:kudo-richardson-2018-sentencepiece}. We report BLEU on the newstest2014 in all the experiments and use newstest2015 for the final evaluation in Section \ref{ssec:combined_results}


All experiments are implemented in fairseq \citep{fairseqott2019}. The configuration of teacher-student training follows the settings in \citet{ludicrously:kim2019}. We use an effective batch size of $458$k words and $16$ GPUs for training. Adam optimizer is applied with $\beta = (0.9, 0.98)$. We use label smoothing with $\varepsilon = 0.1$, inverse square root learning rate schedule with $2500$ warmup steps and peak learning rate of $0.0007$. The models are trained with $50$k updates except for the models with pruning, where additional fine-tuning with $100$-$150$k updates is applied. 
We use a beam size of $5$ during inference. We evaluate the inference speed with batch size of $128$ sentences on GPU, batch size $1$ on CPU and report speed in words per second (wps), averaged over $10$ decoding runs.


\textbf{Hardware:} We evaluate our performance on $1$ GPU (NVIDIA Tesla V100-SXM2-32GB) and $1$ core CPU (Intel Xeon E5-2640 v4 @ 2.40GHz)

\subsection{Replacing Self-Attention with RNN}
\label{ssec:ssru_result}

\begin{table}[h]
    \centering
    \begin{tabular}{lccc}
    \toprule[1.2pt]
         & BLEU & wps & speedup\\
    \midrule[1.2pt]
        Baseline & 28.9 & 4510 & - \\ 
    \midrule[0.5pt]
        AAN & 28.9 & 5323 & 18\% \\ 
        SSRU & 28.7 & 5629 & 25\% \\
    \midrule[0.3pt]
        AAN  w/o ffn & 28.0 & 5915 & 31\% \\ 
        SSRU  w/o ffn & 28.5 & 6079 & 35\% \\
    \bottomrule[1.2pt]
    \end{tabular}
    \caption{\label{table:lightweight} 
    Results of replacing self-attention with lightweight recurrent units and removing the feed-forward network (ffn) in the decoder. Decoding on a GPU with batch-size $128$. 
    }
\end{table}

From Table~\ref{table:lightweight}, we can observe that replacing the self-attention with lightweight recurrent units gives significant speed improvements (18-25\%) without any impact on BLEU score. 

Removing the feed-forward network in the decoder leads to an additional 10-13\% speedup for both AAN and SSRU, but results in $0.9$ BLEU degradation for AAN.
Therefore, we use SSRU as our main architecture in further experiments.

\subsection{Number of Layers}
\label{ssec:layer_result}

We evaluate different combinations of depths in the encoder and decoder. In the decoder, the self-attention mechanism is replaced by the SSRU, and the feed-forward network is removed.

\begin{figure}[ht]
    \centering
    \includegraphics[width=0.5\textwidth]{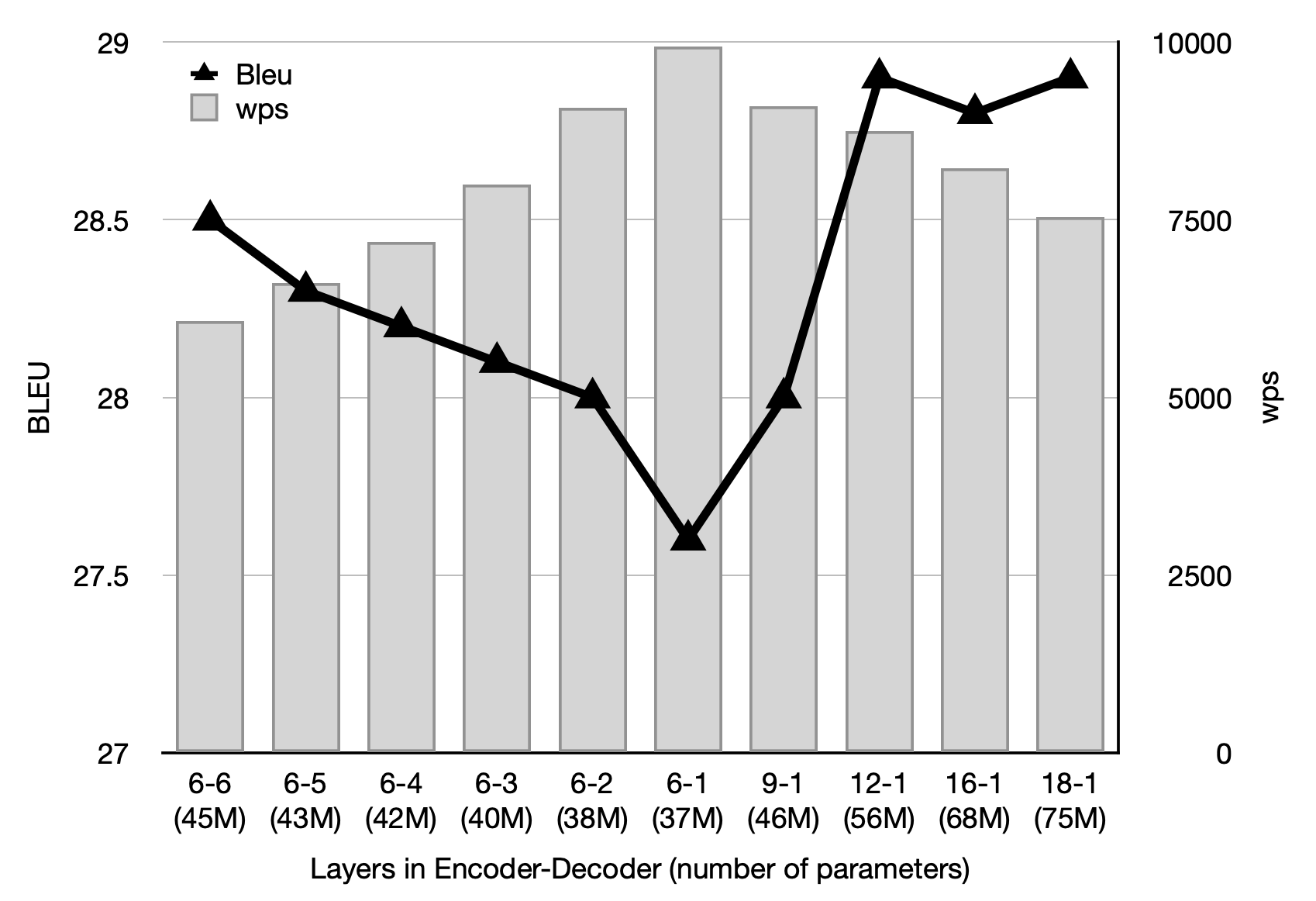}
    \caption{\label{figure:different_layers} Translation quality -- inference speed tradeoff over different number of encoder -- decoder layers.}
\end{figure}
From Figure~\ref{figure:different_layers}, removing one decoder layer at a time from the baseline model increases wps by $10$\% at a cost of BLEU score degradation since model capacity goes down.
As we increase the number of encoder layers to $12$ or more, we observe up to $45$\% speedup, better BLEU score but higher number of parameters than the original $6$-$6$ structure.

\subsection{Pruning Attention Heads} 
\label{ssec:prunehead_results}





\begin{table}[ht]
    \centering
    \begin{tabular}{lcc}
    \hline
    \toprule[1.2pt]
         & attention heads & \multirow{2}{*}{BLEU} \\
         & (enc/enc-dec/dec) \\ 
         \midrule[1.2pt]
        Baseline & 96/8/8 & 29.2 \\
         + pruned & 22/7/8 & 29.0 \\ \hline
        SSRU w/o ffn & 96/8/- & 28.9 \\ 
         + pruned & 18/8/- & 28.6 \\
    \bottomrule[1.2pt]
    \end{tabular}
    \caption{
    Head pruning through $L_0$ regularization on the $12-1$ layer (encoder-decoder) structure. 
    The (enc/enc-dec/dec) refers to the total number of attention heads in encoder self-attention, encoder-decoder attention and decoder self-attention respectively.}
    \label{table:prune_voita}
\end{table}
Pruning allows us to remove up to $75$\% of attention heads with slight BLEU degradation. We observe from the remaining heads that for the pruned baseline ($22$/$7$/$8$) model, the self-attention heads are more important in the deeper layers rather than the lower layers. On the other hand, in our best configuration (SSRU $18$/$8$/-), there is no clear pattern of remaining heads.



\subsection{Combined Results}
\label{ssec:combined_results}
We combine all of the methods and evaluate our model on the newstest2015 testset. 
\begin{table}[ht]
    \centering
    \begin{tabular}{lccc}
    \toprule[1.2pt]
         & \multirow{2}{*}{BLEU} & speedup & \multirow{2}{*}{\#params} \\ 
         & & GPU/CPU & \\
    \midrule[1pt]
        Baseline & 31.1 & - & 61M \\ 
        \hline
        SSRU & 31.1 & 14/12\% & 57M \\ 
        + Remove ffn & 31.0 & 28/49\% & 45M \\ 
        + 12-1 & 31.5 & 82/103\% & 56M \\ 
        + Prune heads & 31.4 & 84/109\% & 46M \\
    \bottomrule[1.2pt]
    \end{tabular}

    \caption{Decoding on a GPU with batch-size $128$, and a single CPU core with batch-size $1$. [$12$-$1$] refers to the number of layers in the encoder and the decoder.}
    \label{table:cpudecode}
\end{table}

Table~\ref{table:cpudecode} shows that by using all of the techniques in combination, the model achieves $84$\% and $109$\% speed improvement on GPU and CPU, respectively compared to the baseline model (Transformer-base). There are only $25$\% heads remain in the deep-encoder after pruning and the total number of parameters is $25$\% fewer. 


\section{Conclusion}
\label{sec:conclusion}

In this paper we explored the combination of techniques aimed at improving inference speed which lead to the discovery of a very efficient architecture. The best architecture has a deep $12$-layer encoder, and a shallow decoder with only one single lightweight recurrent unit layer and one encoder-decoder attention mechanism. $75$\% of the encoder heads were pruned giving rise to a model with $25$\% fewer parameters than the baseline Transformer. In terms of inference speed, the proposed architecture is $84$\% faster on a GPU, and $109$\% faster on a CPU. 


\section*{Acknowledgments}
We would like to thank Andrew Finch, Stephan Peitz, Udhay Nallasamy, Matthias Paulik and Russ Webb for their helpful comments and reviews. Many thanks to the rest of the Machine Translation Team for interesting discussions and support.

\bibliographystyle{acl_natbib}
\bibliography{anthology,emnlp2020}






\end{document}